\documentclass[10pt,twocolumn,letterpaper]{article}

\usepackage{cvpr}
\usepackage{times}
\usepackage{epsfig}
\usepackage{graphicx}
\usepackage{amsmath}
\usepackage{amssymb}

\usepackage{subcaption}
\usepackage{mwe}

\usepackage[pagebackref=true,breaklinks=true,letterpaper=true,colorlinks,bookmarks=false]{hyperref}

\cvprfinalcopy 

\ifcvprfinal\fi
\begin{document}

\title{Multi-modal Ensemble Classification for Generalized Zero Shot Learning}

\author{Rafael Felix \,\,
        Michele Sasdelli \,\,
        Ian Reid \,\,
        Gustavo Carneiro \vspace{3pt} \\
        Australian Institute of Machine Learning \\
        The University of Adelaide \\
        {\tt\small \{rafael.felixalves,michele.sasdelli,ian.reid,gustavo.carneiro\}@adelaide.edu.au}}

\maketitle

\begin{abstract}
Generalized zero shot learning (GZSL) is defined by a training process containing a set of visual samples from seen classes and a set of semantic samples from seen and unseen classes, while the testing process consists of the classification of visual samples from seen and unseen classes.  
Current approaches are based on testing processes that focus on only one of the modalities (visual or semantic), even when the training uses both modalities (mostly for regularizing the training process).  This under-utilization of modalities, particularly during testing, can hinder the classification accuracy of the method.  In addition, we note a scarce attention to the development of learning methods that explicitly optimize a balanced performance of seen and unseen classes.  Such issue is one of the reasons behind the vastly superior classification accuracy of seen classes in GZSL methods.  In this paper, we mitigate these issues by proposing a new GZSL method based on multi-modal training and testing processes, where the optimization explicitly promotes a balanced classification accuracy between seen and unseen classes.  Furthermore, we explore Bayesian inference for the visual and semantic classifiers, which is another novelty of our work in the GZSL framework.
Experiments show that our method holds the state of the art (SOTA) results in terms of harmonic mean (H-mean) classification between seen and unseen classes and area under the seen and unseen curve (AUSUC) on several public GZSL benchmarks.
\end{abstract}


\section{Introduction}

\begin{figure*}[t]
	\centering
	\includegraphics[scale=0.7]{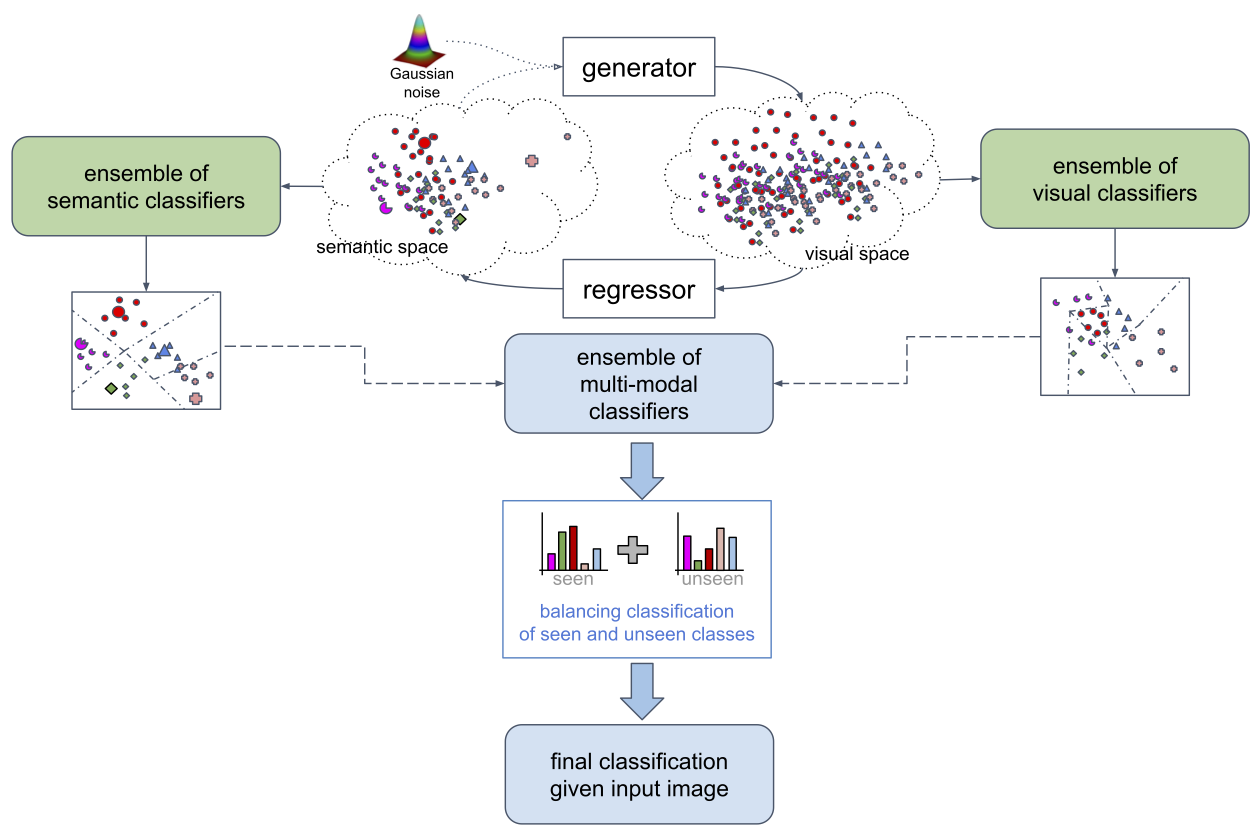}
	\caption{Depiction of the method proposed in this paper -- training and inference processes that explicitly incorporates both modalities (visual and semantic) used in GZSL; a training process that promotes a balanced classification of seen and unseen classes; and a Bayesian classifier, represented by an ensemble classification, for each modality.}
	\label{fig:motivation}
\end{figure*}

As computer vision systems start to be deployed in unstructured environments, it is important that they have the ability to recognize not only the visual classes used during the training process (i.e., the seen classes), but also classes that are not available during training (i.e., unseen classes).  The importance of such ability lies in the impracticality of collecting visual samples from all possible classes that will ever be shown to the system.  In this context, approaches categorized as Generalized Zero-Shot Learning (GZSL)~\cite{chao2016empirical,felix2018multi,xian2017zero} play an important role due to their capacity of classifying visual samples from seen and unseen classes during inference time. 

In general, the training of GZSL methods involves the use of visual samples from seen classes and semantic samples (e.g., textual definition) from seen and unseen classes, where the rationale behind the use of such semantic samples is that, differently from labeled visual samples, they are readily available from various sources, such as Wikipedia, English dictionary~\cite{mikolov2013distributed}, or manually annotated attributes~\cite{lampert2009lerning}.  Such training setup can potentially mitigate the issue of collecting visual samples from all possible unseen classes, but the success of this approach lies in the effective transferring of knowledge between the semantic and visual domains.  Some GZSL methods have focused on a training approach that maps from the visual to semantic samples~\cite{lampert2014attribute}, which are then used in a nearest neighbor classification process.  Other GZSL methods are based on a generative model that takes semantic samples as input to produce visual samples, which are then used for training a classifier of visual samples from seen and unseen classes~\cite{bucher2017generating,felix2018multi,verma2018generalized,xian2017feature}.  There are two interesting aspects worth noting about these GZSL methods~\cite{bucher2017generating,felix2018multi,verma2018generalized,xian2017feature}: 1) even though the training involves the two modalities, 1) they never involve a truly multi-modal classification optimization (a possible exception here is the cycle-consistent multi-modal approach~\cite{bucher2017generating,felix2018multi,verma2018generalized,xian2017feature}, but it does not involve the training of an actual multi-modal classifier); and 2) none of previous approaches rely on a multi-modal inference process.  The under-utilization of both GZSL modalities is problematic because evidence shows that multi-modal ensemble methods generally produce superior classification accuracy, when compared to the results produced by each modality independently~\cite{zhou2012ensemble}.  

Furthermore, even though GZSL methods tend to produce remarkably imbalanced classification accuracy between seen and unseen classes~\cite{chao2016empirical,xian2017zero}, we notice a scarce literature devoted to learning methods that explicitly optimize a balanced classification accuracy between seen and unseen classes~\cite{chao2016empirical}.  A training process that incorporates such balanced optimization is more likely to produce a better harmonic mean (H-mean) classification result between seen and unseen classes.

In this paper, we propose new training and inference methods that explore the multi-modality aspect of GZSL with an ensemble of visual and semantic classifiers~\cite{zhou2012ensemble}.  Furthermore, we explore a Bayesian inference method~\cite{gal2015dropout,gal2017concrete,kendall2017uncertainties} that represents an ensemble of classifiers for each modality (depicted as green color boxes in Fig.~\ref{fig:motivation}), independently.  Finally, we extend the calibrated stacking method~\cite{chao2016empirical} with a learnable parameterized model that maximizes the H-mean result by balancing the contribution between the classification of seen and unseen classes (Fig..~\ref{fig:motivation} depicts our proposal -- see blue colors).  We show the effectiveness of our proposed GZSL approach by ensembling the semantic classifier \emph{direct attribute prediction} (DAP)~\cite{lampert2009lerning} and the visual classifier \emph{multi-modal cycle-consistent method}~\cite{felix2018multi}. In addition, our calibrated stacking method allows the proposed approach to modulate between GZSL and zero-shot learning (ZSL) problem by just adjusting a hyper-parameter (i.e., we do not need to train a ZSL specific model). 

Experimental results on several publicly available datasets commonly used to benchmark GZSL and ZSL methods show that our proposed GZSL produces significant gains in terms of H-mean, ZSL classification accuracy, and the area under the seen and unseen curve (AUSUC)~\cite{chao2016empirical}. Specifically, on CUB~\cite{welinder2010caltech}, we improve the current state of the art (SOTA) H-mean result from \textbf{52.2\%}~\cite{felix2018multi} to \textbf{54.3\%}. On AwA\cite{lampert2014attribute}, we achieve H-mean of \textbf{60.2\%} compared to \textbf{59.7\%}~\cite{felix2018multi}. On FLO~\cite{nilsback2008automated} and SUN~\cite{xiao2010sun}, we improve the current SOTA H-mean result from \textbf{64.5\%} and \textbf{39.2\%}~\cite{felix2018multi} to \textbf{66.8\%} and \textbf{39.4\%}, respectively. Furthermore, our proposed approach achieves outstanding improvements in terms of ZSL accuracy for CUB and FLO, from \textbf{57.8\%} and \textbf{68.8\%}~\cite{felix2018multi} to \textbf{71.1\%} and \textbf{83.9\%}, respectively (note that this is achieved by modulating the calibrated stacking method, as decribed above, i.e., without re-trained our GZSL model). Finally, in addition to such results, we assess our approach in terms of AUSUC~\cite{chao2016empirical}, which allows us to provide more thorough comparisons for the GZSL problem.  In particular, our model
shows an AUSUC of \textbf{0.430, 0.673, 0.477} for the datasets CUB, FLO and AWA, while the SOTA~\cite{felix2018multi} produces \textbf{0.418, 0.595, 0.473} on the same respective datasets.


\section{Literature Review}
\label{sec:literature_review}

In this section we present relevant literature that contextualizes and motivates our approach. 

\textbf{Generalized Zero-Shot Learning (GZSL).} In recent years, we have observed a growing interest in GZSL. Part of the reason behind such interest derives from the paper by Xian et al.~\cite{xian2017zero} that clearly formalizes the GZSL problem and introduces a more solid experimental setup and a new evaluation metric based on the H-mean between the classification accuracy results of the seen and the unseen visual classes.

Recently proposed GZSL methods can be roughly divided into two categories: \textbf{semantic attribute prediction} and \textbf{visual data augmentation}. \textbf{Semantic attribute prediction} methods~\cite{akata2016label,he2015delving,lampert2009lerning} tackle GZSL by training a regressor that maps visual samples from seen classes to their respective semantic samples. Therefore, given a new test visual sample (from a seen or unseen class), the regressor maps it into a semantic sample that is then used in a nearest neighbor classification process. The main assumption explored in this approach is that the mapping from the visual to the semantic domain learned for the seen classes can be transferred into the unseen classes. Unfortunately, such assumption is unwarranted, and the main issue affecting this approach is that the test visual samples from seen classes are often classified correctly, but the ones from the unseen classes are usually classified incorrectly into one of the seen classes~\cite{xian2017zero}.  

\textbf{Visual data augmentation} relies on a generative model that is trained to produce visual samples from their respective semantic samples~\cite{bucher2017generating,felix2018multi,verma2018generalized,xian2017feature}. Such model allows the generation of visual samples from unseen classes, which are then used in the modelling of a classifier that is trained with real visual samples from seen classes and generated samples from unseen classes.  Methods based on this approach hold the SOTA GZSL results~\cite{bucher2017generating,felix2018multi,verma2018generalized,xian2017feature}.  Recently, this approach has been extended in its training process, where the generated visual samples are forced to regress correctly to their respective semantic samples, forming a multi-modal cycle consistent training~\cite{felix2018multi,verma2018generalized}.  This extension represents the first attempt at a multi-modal training, which allowed a further improvement to the current SOTA.  It is interesting that the inference process of \textbf{semantic attribute prediction} focuses exclusively on the semantic space, while \textbf{visual data augmentation} works solely on the visual space.  Therefore, a natural question is: can an ensemble of semantic and visual classifiers complement each other, particularly during inference?  This is one of the questions we aim to answer with this paper.

\textbf{Ensemble Classification.} In machine learning and pattern recognition, several studies confirm the effectiveness of combining classifiers to enhance performance~\cite{dietterich2000ensemble,dietterich2000experimental,zhou2012ensemble}, where the diversity of the models and the way that they complement each other are important factors that affect the ensemble classification effectiveness.
An ensemble of classifiers consists of approaches that combine the decisions made by multiple models~\cite{dietterich2000ensemble} -- we highlight two specific methods that are explored in this work: \textbf{multi-modal ensemble} and \textbf{Bayesian weighted voting}. \textbf{Multi-modal ensemble} methods 
combine the classification performed using multiple modalities of the input data. 
\textbf{Bayesian weighted voting} approaches consist of averaging the output probability from different models applied to the same input data.  In particular, we explore the method introduced by Gal and Ghahramani~\cite{gal2015dropout}, who formulated an approximated Bayesian Neural Networks with the use of dropout variational inference. In this formulation, dropout~\cite{srivastava2014dropout}, which is often used as a training regularizer, is run at inference time to estimate the mean and variance of model predictions. 

To the best of our knowledge, no attention has been devoted to the design and implementation of any type of ensemble classification in GZSL. In this paper, we propose a new GZSL approach relying on the two types of ensemble classification highlighted above. 

\textbf{Balancing the Contribution of Seen and Unseen Classifiers.}  The work by Chao et al.~\cite{chao2016empirical} introduces an important point in the GZSL problem, which is the clear imbalance in terms of the classification accuracy between the seen and unseen classes.  Even though Chao et al.'s paper~\cite{chao2016empirical} has received relatively sparse attention, we agree that this is a crucial point that must be explored further. We note that, although H-mean has been extensively used to evaluate GZSL methods~\cite{long2018pseudo,xian2017zero}, it is unclear whether previous approaches have tried to optimize H-mean during the training of a GZSL classifier. We argue that training GZSL to maximize H-mean can mitigate the imbalanced classification of GZSL approaches, resulting in more balanced classification results for seen and unseen classes.

\section{Method}
\label{sec:method}

In this section we introduce the problem formulation and our proposed approach. 

\subsection{Problem Formulation}
\label{sec:problem_formulation}

In order to formulate the GZSL problem~\cite{chao2016empirical,xian2017zero}, we first define the visual dataset $\mathcal{D} = \{(\textbf{x},  y)_i\}_{i=1}^{N}$, where $\textbf{x} \in \mathcal{X} \subseteq \mathbb{R}^K$ denotes the visual representation
(acquired from the second to last layer of a pre-trained deep residual nets~\cite{he2016resnet}), and $y \in \mathcal{Y} = \{ 1,..., C \}$ denotes the visual class. 
The visual dataset has $N$ samples, denoting the number of images. We also need to define the semantic dataset $\mathcal{R} = \{ (\mathbf{a},y)_j \}_{j \in \mathcal{Y}}$, which associates visual classes with semantic samples, where $\mathbf{a} \in \mathcal{A} \subseteq \mathbb R^L$ represents a semantic feature (e.g., set of continuous features as \textit{word2vec}~\cite{xian2017zero}). The semantic dataset has as many elements as the number of classes. The set $\mathcal{Y}$ is split into the seen subset $\mathcal{Y}^S = \{1, ... , S\}$, and the unseen subset $\mathcal{Y}^U = \{(S+1), ... , (S+U)\}$. Therefore, $C=S + U$, with $\mathcal{Y} = \mathcal{Y}^S \cup \mathcal{Y}^U$, $\mathcal{Y}^S \cap \mathcal{Y}^U = \emptyset$.
Furthermore, $\mathcal{D}$ is also divided into mutually exclusive training and testing visual subsets $\mathcal{D}^{Tr}$ and $\mathcal{D}^{Te}$, respectively,
where $\mathcal{D}^{Tr}$ contains a subset of the visual samples belonging to the seen classes, and $\mathcal{D}^{Te}$ has the visual samples from the seen classes held out from training and all samples from the unseen classes.  
The training dataset is composed of the semantic dataset $\mathcal{R}$ and the training visual subset $\mathcal{D}^{Tr}$, while the testing dataset relies only on the testing visual subset $\mathcal{D}^{Te}$.

In the next sub-sections, we first define the \textbf{semantic attribute prediction} and the \textbf{visual data augmentation} methods introduced in Sec.~\ref{sec:literature_review}, where we also introduce the ensemble classification for each method.  Then, we define the multi-modal GZSL ensemble classifier and the optimization of a balanced classification between seen and unseen classes.

\subsection{Semantic Attribute Prediction}
\label{sec:dap}

As explained in Sec.~\ref{sec:literature_review}, semantic attribute prediction methods consist of a regressor that are trained to map visual samples from seen classes to their respective semantic samples.  In particular, we explore the Direct Attribute Prediction (DAP) formulation~\cite{lampert2009lerning}, defined by:
\begin{equation}
    {\textbf{a}^*} = r(\textbf{x}; \omega_t ),
    \label{eq:dap}
\end{equation}
where $r(.)$ represents a regressor parameterized by $\omega_t \in \Omega$, and $t$ indexes a particular model belonging to an ensemble of models, as defined below.
In DAP, this regressor is represented by a linear transform trained by minimising the mean square error (MSE) of the reconstructed semantic sample in $\mathcal{R}$ using the visual samples in $\mathcal{D}^{Tr}$.
In order to build the ensemble of semantic classifiers, we resort to Monte Carlo (MC) dropout inference, consisting of $T$ forward passes of a deep learning model (each pass having a different dropout map), where each pass represents an approximate MC sampling (note that MC Dropout can be seen as a way of approximating an ensemble of $T$ models)~\cite{gal2015dropout}. This relies on training and testing the regressor in \eqref{eq:dap} with dropout. Effectively, this means that the regressor in \eqref{eq:dap} is in fact parameterized by $\omega_t$, where $t \in \{1,...,T\}$ indexes a unique dropout inference that can produce a slightly different regression result.  The DAP nearest neighbor classification probability is achieved as follows:
\begin{equation}
p_R(y^* | \textbf{x},\omega_t) = \begin{cases}
1, \text{if } y^* = \underset{(\textbf{a}, y) \in \mathcal{R}}{\arg\min}  \Big|\Big|  r(\textbf{x}; \omega_{t}) - \textbf{a} \Big|\Big|^{2},\\
0, \text{otherwise}.
\end{cases}
\label{eq:score-1nn}
\end{equation}
The MC dropout inference~\cite{gal2015dropout} consists of averaging the classification results:
\begin{equation}
p_{dap}(y | \textbf{x}, \omega) = \frac{1}{T} \sum_{t=1}^{T} p_R ( y |\textbf{x}; \omega_t ),
\label{eq:score-dap}
\end{equation}
where $p_R(.)$ represents the output probability of the one forward pass defined in \eqref{eq:score-1nn}, and $p_{dap}(.)$ denotes the classification probability from the MC Dropout ensemble.

\subsection{Visual Data Augmentation}
\label{sec:cyg}

This type of GZSL has been introduced in Sec.~\ref{sec:literature_review}, where the main idea is the use of a generative model~\cite{xian2017feature} that is trained to randomly produce visual samples conditioned on their semantic samples.  After training this generative model, it is then possible to generate visual samples from the unseen classes in order to train a classifier using actual visual samples from the seen classes and generated visual samples from the unseen classes~\cite{bucher2017generating,felix2018multi,verma2018generalized,xian2017feature}.  This approach has been recently extended with a cycle consistency loss that trains this classifier, together with the regressor in \eqref{eq:dap} in order to regularize the training process -- this approach is referred to as cycle-WGAN~\cite{felix2018multi}, and the classification produced for new visual test samples is defined by:
\begin{equation}
    y^* = \arg\max_{y \in \mathcal{Y}} p_D( y  | \textbf{x}, \theta_t),
\label{eq:wgan_cycle}
\end{equation}
where $p_D(.)$ is the visual classifier parameterized by $\theta_t \in \Theta$. Similarly to the DAP classification defined in \eqref{eq:score-dap}, we train the classifier in \eqref{eq:wgan_cycle} using dropout \cite{gal2015dropout}, and the testing is indexed by $t \in \{1,...,T\}$, representing a unique dropout inference.  Hence, the classification probability of a test visual sample produced by the cycle-WGAN is defined as:
\begin{equation}
    p_{cyg}(y | \textbf{x}, \theta) = \frac{1}{T} \sum_{t=1}^{T} p_D( y | \textbf{x}; \theta_t  ).
\label{eq:score-cyc}
\end{equation}

\subsection{Multi-Modal Ensemble of GZSL classifiers}
\label{sec:multimoda}

Our proposed multi-modal ensemble classification combines the semantic attribute prediction classifier in \eqref{eq:score-dap} and the visual data augmentation classifier in \eqref{eq:score-cyc}.  The idea is to formulate a parameterized ensemble classification based on agreement voting, if both classifiers agree on the same class that maximises the classification in the respective space.  However, if the classifiers disagree, then we use a weighted average of these classifiers, as follows:
\begin{equation}
\begin{split}
p_{ens}&(y^* | \textbf{x},\omega,\theta,\alpha) = \\
&\begin{cases}
1,  \text{if } y^*= \underset{y \in \mathcal{Y}}{\arg\max}  p_{dap}(y|\mathbf{x},\omega) \text{ and } \\
  \;\;\;\;\;\; y^*= \underset{y \in \mathcal{Y}}{\arg\max} p_{cyg}(y|\mathbf{x},\theta); \\
 \alpha \times p_{cyg}(y^* | \textbf{x}, \theta) + 
 (1-\alpha) \times p_{dap}(y^* |\textbf{x}, \omega) , \text{ o.w.},
\end{cases}
\end{split}
\label{eq:combine scores}
\end{equation}
where $\alpha \in [0,1]$ denotes the ensemble classification parameter.  The optimization of $\alpha$ in \eqref{eq:combine scores} is performed by first splitting the original training set $\mathcal{D}^{Tr}$ into two subsets: a (sub-)training set $\widehat{\mathcal{D}}^{Tr} \subset \mathcal{D}^{Tr}$ containing a subset of the original seen classes $\widehat{\mathcal{Y}}^S \subset \mathcal{Y}^S$ and a validation set $\mathcal{D}^{Va} \subset \mathcal{D}^{Tr}$ with the remaining unseen classes $\widehat{\mathcal{Y}}^U \subset \mathcal{Y}^S$. Then, $\alpha$  is estimated with:
%
%
\begin{equation}
\footnotesize{
\alpha^* = \arg\max_{\alpha} H(Acc(p_{ens}( .|.,\alpha),\widehat{\mathcal{Y}}^S),Acc(p_{ens}( .|.,\alpha),\widehat{\mathcal{Y}}^U)),
}
\label{eq:alpha}
\end{equation}
where: 
\begin{equation}
\begin{split}
    H( Acc(p(.) & ,\widehat{\mathcal{Y}}^S), Acc(p(.),\widehat{\mathcal{Y}}^U)) = \\
    & 2 \times \frac{\big( Acc(p(.), \widehat{\mathcal{Y}}^{S}) \times Acc(p(.), \widehat{\mathcal{Y}}^{U}) \big)} {\big( Acc(p(.), \widehat{\mathcal{Y}}^{S}) + Acc(p(.), \widehat{\mathcal{Y}}^{U}) \big)}
    \label{eq:hmean}
\end{split}
\end{equation}
%
is the harmonic mean between the seen and unseen accuracies
with $Acc(p(.),\widehat{\mathcal{Y}})$ denoting the accuracy of the classifier $p(.)$ on the set $\widehat{\mathcal{Y}}$.


\subsection{Classification Weighting of Seen and Unseen Classes}
\label{sec:balance}

Another contribution of this paper is a simple parameterized weighting method that balances the classification of seen and unseen classes, which is defined as:
\begin{equation}
\begin{split}
b(y ; & \textbf{x},  \mathcal{Y}^U, \mathcal{Y}^S, \phi, \beta ) = \\
& p(y | \textbf{x}, \phi) \left ( \beta\times{\mathbb I(y \in \mathcal{Y}^U)} + (1-\beta)\times{\mathbb I(y \in \mathcal{Y}^S)} \right ),
\end{split}
\label{eq:score-by-beta}
\end{equation}
where $\beta \in [0,1]$ represents the learnable parameter of this class weighting function, $\phi$ is the parameter of the classifier $p(.)$, and
$\mathbb I(.)$ denotes an indicator function that is 1 if the condition represented in its parameter is true, and 0 otherwise.
The optimization of $\beta$ in \eqref{eq:score-by-beta} is performed in the same way as the optimization of $\alpha$ in \eqref{eq:alpha}.  Note that in \eqref{eq:score-by-beta} , we do not refer to any of the previously defined classifiers because such weighting can in fact be applied to any one of them. 

\section{Experiments}
\label{sec:experiments}

In this section we introduce the benchmark datasets, as well as the evaluation criteria. Then, we discuss the setup adopted for the experiments, present the results of our proposal, and we finally compare them with the current SOTA.

\subsection{Datasets}
\label{sec:datasets}

We assess our method on publicly available benchmark GZSL datasets. More specifically, we perform experiments on CUB-200-2011~\cite{welinder2010caltech,xian2017zero}, FLO~\cite{nilsback2008automated}, SUN~\cite{xian2017zero}, and AWA~\cite{lampert2009lerning,xian2017zero} with the GZSL experimental setup described by Xian et al.~\cite{xian2017zero}. The datasets CUB and FLO are generally regarded as fine-grained, while AWA and SUN consist of coarse datasets. In Table~\ref{table:dataset-stats} we display some basic information about the datasets in terms of number of seen and unseen classes and number of training and testing images.

\begin{table*}[t]
\centering
\caption{Information about the datasets CUB\cite{welinder2010caltech}, FLO\cite{nilsback2008automated}, SUN \cite{xiao2010sun}, AWA\cite{xian2017zero}, and ImageNet~\cite{deng2009imagenet}. Column (1) shows the number of seen classes, denoted by $|\mathcal{Y}^S|$, split into the number of training and validation classes (train+val), (2) presents the number of unseen classes $| \mathcal{Y}^U |$, (3) displays the number of samples available for training $|\mathcal{D}^{Tr}|$ and (4) shows number of testing samples that belong to the unseen classes $|\mathcal{D}_U^{Te}|$ and number of testing samples that belong to the seen classes $|\mathcal{D}_S^{Te}|$ from~\cite{felix2018multi}.}
\label{table:dataset-stats}
\begin{tabular}{|l|c|c|c|c|}
\hline
\textbf{Name} & $|\mathcal{Y}^S|$ (train+val) & $|\mathcal{Y}^U|$ & $|\mathcal{D}^{Tr}|$ & $|\mathcal{D}^{Te}_U|+|\mathcal{D}^{Te}_S|$
\\
\hline
\hline
CUB     & 150 (100+50) & 50 & 7057 & 1764+2967 \\
FLO     & 82 (62+20) & 20 & 1640 & 1155+5394 \\
SUN    & 745 (580+65) & 72 & 14340 & 2580+1440\\
AWA~\footnote{Dataset from https://cvml.ist.ac.at/AwA2/.}    & 40 (27+13) & 10 & 19832 & 4958+5685\\
ImageNet & 1000 (1000 + 0) & 100 & $1.2$kk & 5200+50k \\
\hline
\end{tabular}
\end{table*}

For the semantic samples, we use the 1024-dimensional vector produced by CNN-RNN~\cite{reed2016learning} for CUB-200-2011~\cite{xian2017zero} and FLO~\cite{nilsback2008automated}. These semantic samples represent the textual description of each image using 10 sentences per image. In order to define a unique semantic sample per-class, we average the semantic samples of all images belonging to each class~\cite{xian2017zero}. For the SUN~\cite{xian2017zero} and AWA~\cite{xian2017zero} datasets, we use manually annotated semantic samples containing respectively 102 and 85 dimensions. For the visual samples, we follow the protocol by Xian et al.~\cite{xian2017zero}, where the features are represented by the activations of the 2048-dimensional top pooling layer of ResNet-101~\cite{he2016resnet}, obtained for the image.

For ImageNet~\cite{deng2009imagenet}, there are several testing splits for GZSL (e.g., 2-hop, 3-hop), which rely on the training set of 1K classes and testing set on 22K classes. However, recent studies reported that there is overlap between seen and unseen classes for GZSL~\cite{felix2018multi}. We argue that although these splits may be suitable for Open Set Recognition approaches, further studies are required to reassure their applicability for GZSL~\cite{felix2018multi}. Nevertheless, in order to demonstrate the robustness of our approach to large datasets, we perform an experiment with ImageNet~\cite{deng2009imagenet} for a split containing 100 classes for testing ~\cite{wang2017multi} and the standard 1K classes for training \cite{wang2017multi}, without overlap between seen and unseen classes. For ImageNet, we used $500$-dimensional semantic samples~\cite{wang2017multi} and $2048$-dimensional ResNet-features, where images are resized to $256 \times 256$ pixels, cropped to $224 \times 224$ pixels,  normalized with means (0.485, 0.456, 0.406) and standard deviations (0.229, 0.224, 0.225) per RGB channel.

\subsection{Evaluation Protocol}
\label{sec:evaluation_protocol}

The evaluation protocol is based on computing the average per-class top-1 accuracy measured independently for each class before dividing their cumulative sum by the number of classes~\cite{xian2017zero}.
For GZSL, after computing the average per-class top-1 accuracy on seen classes $\mathcal{Y}^S$ and unseen classes $\mathcal{Y}^U$, we compute the H-mean~\eqref{eq:hmean} of seen and unseen classification accuracies~\cite{xian2017zero}.  For ZSL, we only compute the average per-class top-1 accuracy on the unseen classes.  We also show results using a graph that measures the classification accuracy on the seen and unseen classes by varying $\alpha$ in~\eqref{eq:alpha} and $\beta$ in~\eqref{eq:score-by-beta}.  Using such graph, we can optimize the values of $\alpha$ and $\beta$ using a validation set, and we can measure a more general classification accuracy of the GZSL method without commiting to a particular operating point (represented by a specific value for $\alpha$ and $\beta$) -- such general accuracy is represented by the AUSUC~\cite{chao2016empirical}, obtained by calculating the area under the curve that represents the supremum of all points measuring the classification accuracy on the seen and unseen classes for different values of $\beta$.

\subsection{Implementation Details}
\label{sec:implementation_details}

\begin{figure}[h!]
    \centering
    \begin{subfigure}[b]{0.235\textwidth}
        \centering
        \includegraphics[width=\textwidth]{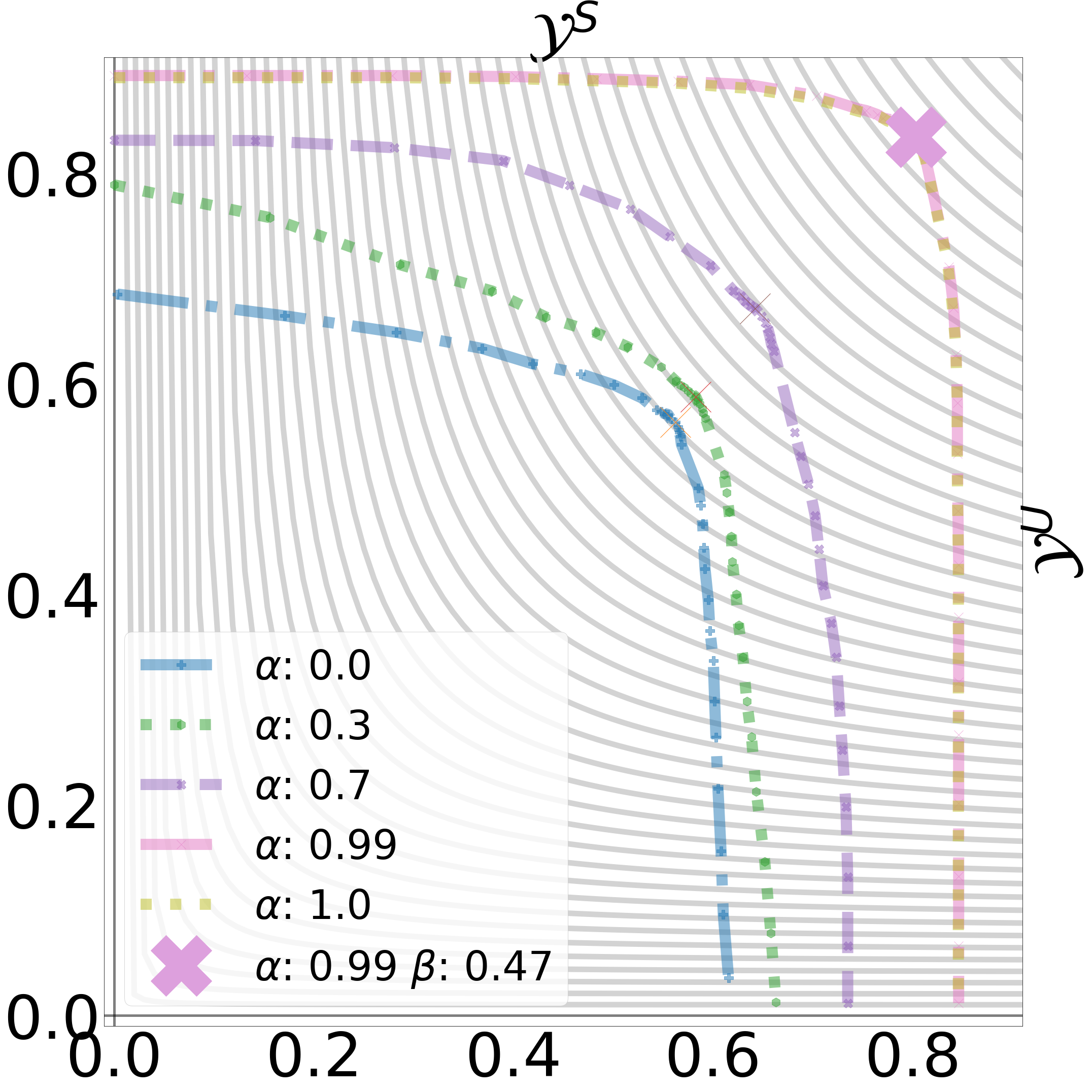}
        \caption{~CUB}
        \label{fig:hp-cub}
    \end{subfigure}
    \begin{subfigure}[b]{0.235\textwidth}
       \centering
        \includegraphics[width=\textwidth]{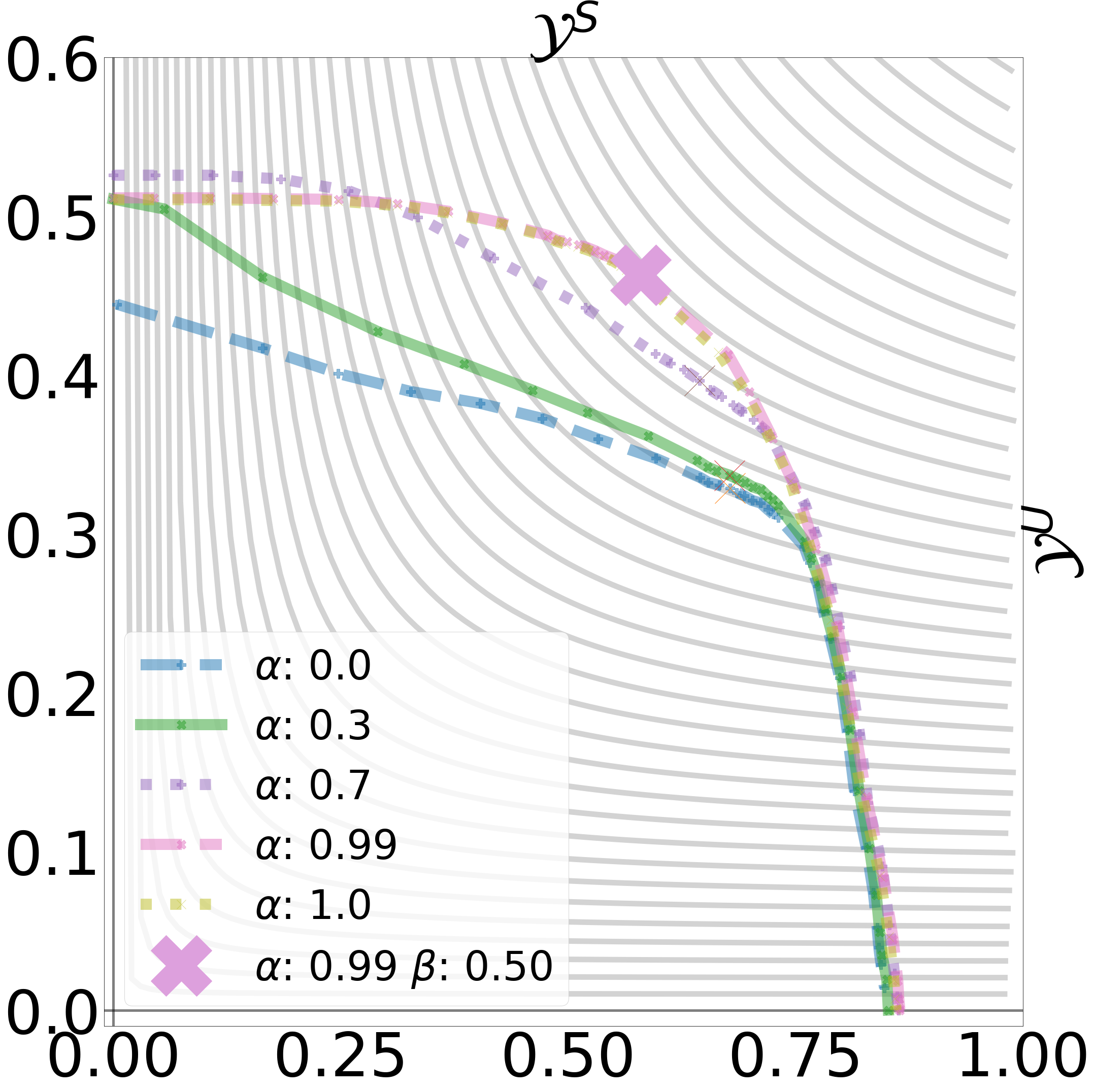}
        \caption{~AWA}
        \label{fig:hp-awa1}
    \end{subfigure}
    \caption{Classification accuracy of model 3ME on seen classes $\mathcal{Y}^S$ (hor. axis) and unseen classes $\mathcal{Y}^U$ (ver. axis).  This graph is used for computing the AUSUC and for optimizing the hyper-parameters $\alpha$ in \eqref{eq:alpha} and $\beta$ in \eqref{eq:score-by-beta} for datasets (a) \textbf{CUB} and (b) \textbf{AWA} on the validation set~\cite{xian2017zero}. Each colored curve represents the seen and unseen classifications for a particular value of $\alpha$, where the variation of $\beta$ produces different points along these $\alpha$-curves; while the grey contours represent H-mean contour lines.  The highlighted point with a pink cross represents the optimal H-mean on the validation set, where the legend shows its respective $\alpha$ and $\beta$ values.  Finally, the AUSUC is computed as the area under one of the $\alpha-$ curves.}
    \label{fig:hp-search}
\end{figure}

In this section, we describe the implementation details for \textbf{DAP}~\cite{lampert2009lerning} (which is a good representative method for the semantic attribute prediction from Sec.~\ref{sec:dap}), \textbf{cycle-WGAN}~\cite{felix2018multi} (good representative method for the visual data augmentation from Sec.~\ref{sec:cyg}), \textbf{MC Dropout}~\cite{gal2015dropout} in \eqref{eq:combine scores} and the hyper-parameters of the ensemble and classification weighting in \eqref{eq:combine scores} and \eqref{eq:score-by-beta}, respectively. In terms of model architecture and hyper-parameters (e.g. \textit{number of epochs, batch size, number of layers, learning rate, and weight decay, learning rate decay}), we followed the configuration by Lampert et al.~\cite{lampert2009lerning} for \textbf{DAP} and by Felix et al.~\cite{felix2018multi} for \textbf{cycle-WGAN}. In terms of MC dropout, we introduce a dropout layer at the input of both networks, with a dropout rate of $0.2$ and a number of forward passes $T = 100$, for both models. In terms of $(\alpha,\beta)$ in~\eqref{eq:alpha} and~\eqref{eq:score-by-beta}, we use the following values: $(0.99, 0.47)$ for CUB and FLO, $(0.99, 0.5)$ for AWA, $(0.99, 0.49)$ for SUN, and $(1.0, 0.51)$ for ImageNet.
Furthermore, we refer to our proposed Monte Carlo Multi-Modal Ensemble as 3ME and the Multi-Modal Ensemble as 2ME (i.e., 2ME is the 3ME model without MC dropout), where both models are formed by the ensemble of our two baseline models DAP \cite{lampert2009lerning} and cycle-WGAN \cite{felix2018multi}.
For all the hyper-parameters above, we estimate their optimal values using a grid search optimization on the validation set proposed in~\cite{xian2017zero}.  
However, it is interesting to show in more detail the optimization of $\alpha$ and $\beta$, depicted in Fig.~\ref{fig:hp-search}. We vary $\alpha \in [0,1]$ and $\beta \in [0,1]$ -- for $\alpha=0$, only the DAP classifier is ``switched on", resulting in a weaker result as depicted in Fig.~\ref{fig:hp-search}. On the other hand, for $\alpha=1$, only cycle-WGAN is ``switched on", producing a stronger result. 
In fact, according to this figure, an increase of $\alpha$ generally allows an increase of the AUSUC (measured as the area under each of the curves, represented by a specific $\alpha$ value). 
Also, the variation of $\beta$ moves the results from a ZSL classifier when $\beta=1$ (top-left part of the curve) to a fully supervised classifier when $\beta=0$ (bottom-right part of the curve).  Hence, a value of $\beta$ around $0.5$ generally means a balanced classification between seen and unseen classes, resulting in an optimal H-mean value.

\subsection{Results}

In this section we show the GZSL results using our proposed models 2ME and 3ME and several other baseline methods previously used in the field for benchmarking~\cite{felix2018multi,xian2017zero,xian2017feature}.  Table~\ref{table:gzsl_results} shows the benchmark GZSL results on CUB, FLO, SUN and AWA using the evaluation measures mentioned in Sec.~\ref{sec:evaluation_protocol}, where $\mathcal{Y}^U$, $\mathcal{Y}^S$ and $H$ represent the GZSL classification accuracy on the unseen test set, seen test set, and H-mean result; while $ZSL$ denotes the ZSL classification accuracy on the unseen test set.
The results in Table~\ref{table:gzsl_imagenet} show the top-1 accuracy on ImageNet for our proposed approach 3ME and the current SOTA.

\begin{table*}[h!]
\centering
\caption{GZSL results using per-class average top-1 accuracy on the test sets of unseen classes $\mathcal{Y}^U$, seen classes $\mathcal{Y}^S$, and H-mean result $H$; and ZSL results on the unseen classes exclusively -- all results shown in percentage. The results from previously proposed methods in the field were extracted from~\cite{XianCVPR2017}.  The highlighted values represent the best ones in each column.}
\label{table:gzsl_results}
\centering

\resizebox{\textwidth}{!}{
\begin{tabular}{|l|llll|llll|llll|llll|}
\hline
& & \textbf{CUB} & & 
& & \textbf{FLO} & &
& & \textbf{SUN} & &
& & \textbf{AWA} & &
\\
\textbf{Classifier}  
&  $\mathcal{Y}^U$ & $\mathcal{Y}^S$ & $H$ & $ZSL$
&  $\mathcal{Y}^U$ & $\mathcal{Y}^S$ & $H$ & $ZSL$
&  $\mathcal{Y}^U$ & $\mathcal{Y}^S$ & $H$ & $ZSL$
&  $\mathcal{Y}^U$ & $\mathcal{Y}^S$ & $H$ & $ZSL$
\\ \hline \hline

DAP~\cite{lampert2009lerning}  
& $ 4.2$ & $25.1$ & $ 7.2$ & $-$
& $ -  $ & $ -  $ & $  - $ & $-$
& $ 1.7$ & $67.9$ & $ 3.3$ & $-$
& $ 0.0$ & $\textbf{88.7}$ & $ 0.0$ & $-$
\\

IAP~\cite{lampert2009lerning}  
& $ 1.0$ & $37.8$ & $ 1.8$ & $-$
& $ -  $ & $ -  $ & $  - $ & $-$
& $ 0.2$ & $\textbf{72.8}$ & $ 0.4$ & $-$
& $ 2.1$ & $78.2$ & $ 4.1$ & $-$
\\

DEVISE~\cite{frome2013devise}
& $23.8$ & $53.0$ & $32.8$ & $52.0$
& $ 9.9$ & $44.2$ & $16.2$ & $45.9$
& $16.9$ & $27.4$ & $20.9$ & $56.5$
& $13.4$ & $68.7$ & $22.4$ & $54.2$
\\

SJE~\cite{akata2015evaluation}        
& $23.5$ & $59.2$ & $33.6$ & $53.9$
& $13.9$ & $47.6$ & $21.5$ & $53.4$
& $14.7$ & $30.5$ & $19.8$ & $53.7$
& $11.3$ & $74.6$ & $19.6$& $\textbf{65.6}$
\\

LATEM~\cite{xian2016latent}      
& $15.2$ & $57.3$ & $24.0$ & $49.3$
& $ 6.6$ & $47.6$ & $11.5$ & $40.4$
& $14.7$ & $28.8$ & $19.5$ & $55.3$
& $ 7.3$ & $71.7$ & $13.3$ & $55.1$
\\

ESZSL~\cite{romera2015embarrassingly}      
& $12.6$ & $\textbf{63.8}$ & $21.0$ & $53.9$
& $11.4$ & $56.8$ & $19.0$ & $51.0$
& $11.0$ & $27.9$ & $15.8$ & $54.5$
& $ 6.6$ & $75.6$ & $12.1$ & $58.2$
\\
ALE~\cite{akata2016label}        
& $23.7$ & $62.8$ & $34.4$ & $54.9$
& $13.3$ & $61.6$ & $21.9$ & $48.5$
& $21.8$ & $33.1$ & $26.3$ & $58.1$
& $16.8$ & $76.1$ & $27.5$ & $59.9$
\\
SAE~\cite{kodirov2017semantic}        
& $ 8.8$ & $18.0$ & $11.8$ & $-$
& $ -  $ & $ -  $ & $  - $ & $-$
& $ 7.8$ & $54.0$ & $13.6$ & $-$
& $ 1.8$ & $77.1$ & $ 3.5$ & $-$
\\

{f-CLSWGAN} \cite{XianCVPR2018} 
& $43.8$ & $60.6$ & $50.8$ & $57.7$ 
& $58.8$ & $70.0$ & $63.9$ & $66.8$
& $47.9$ & $32.4$ & $38.7$ & $58.5$
& $56.0$  & $62.8$ & $59.2$ & $64.1$
\\

{cycle-WGAN} \cite{felix2018multi}
& $46.0$ & $60.3$ & $52.2$ & $57.8$ 
& $\textbf{59.1}$ & $71.1$ & $64.5$ & $68.8$ 
& $\textbf{48.3}$ & $33.1$ & $39.2$ & $\textbf{59.7}$ 
& $\textbf{56.4}$ & $63.5$ & $59.7$ & $\textbf{65.6}$ 
\\ 

\hline

{3ME} 
& $\textbf{49.6}$ & $60.1$ & $\mathbf{54.3}$ & $\textbf{71.1}$ 
& $57.8$ & $\mathbf{79.2}$ & $\textbf{66.8}$ & $\textbf{83.9}$ 
& $44.0$ & $35.8$ & $\textbf{39.4}$ & $58.1$ 
& ${55.5}$ & $65.7$ & $\textbf{60.2}$ & $62.8$ 
\\ \hline
 
\end{tabular}
}
\end{table*}


\begin{table}[h]
\centering
\caption{GZSL and ZSL ImageNet results -- all results shown in percentage.  Please see caption of Table~\ref{table:gzsl_results} for details on each measure.  The highlighted values represent the best ones in each column.}
\label{table:gzsl_imagenet}
\centering
\begin{tabular}{|l|cccc|}
\hline
\textbf{Classifier}  &  $\mathcal{Y}^U$ & $\mathcal{Y}^S$ & $H$ & $ZSL$ \\ 
\hline
\hline
{f-CLSWGAN}~\cite{XianCVPR2018} & $0.7$  & $-$ & $-$ & $7.5$  \\
{cycle-WGAN}~\cite{felix2018multi}  & $1.5$ & $\mathbf{66.5}$ & $2.8$ & $\mathbf{8.7}$ \\
\hline
{3ME}  & $\mathbf{2.5}$ & $47.4$ & $\mathbf{4.8}$ & $7.0$ \\
\hline
\end{tabular}
\end{table}

\begin{figure*}[h!]
  \centering
  \begin{subfigure}[b]{0.245\textwidth}
    \centering
    \includegraphics[width=\textwidth, height=\textwidth]{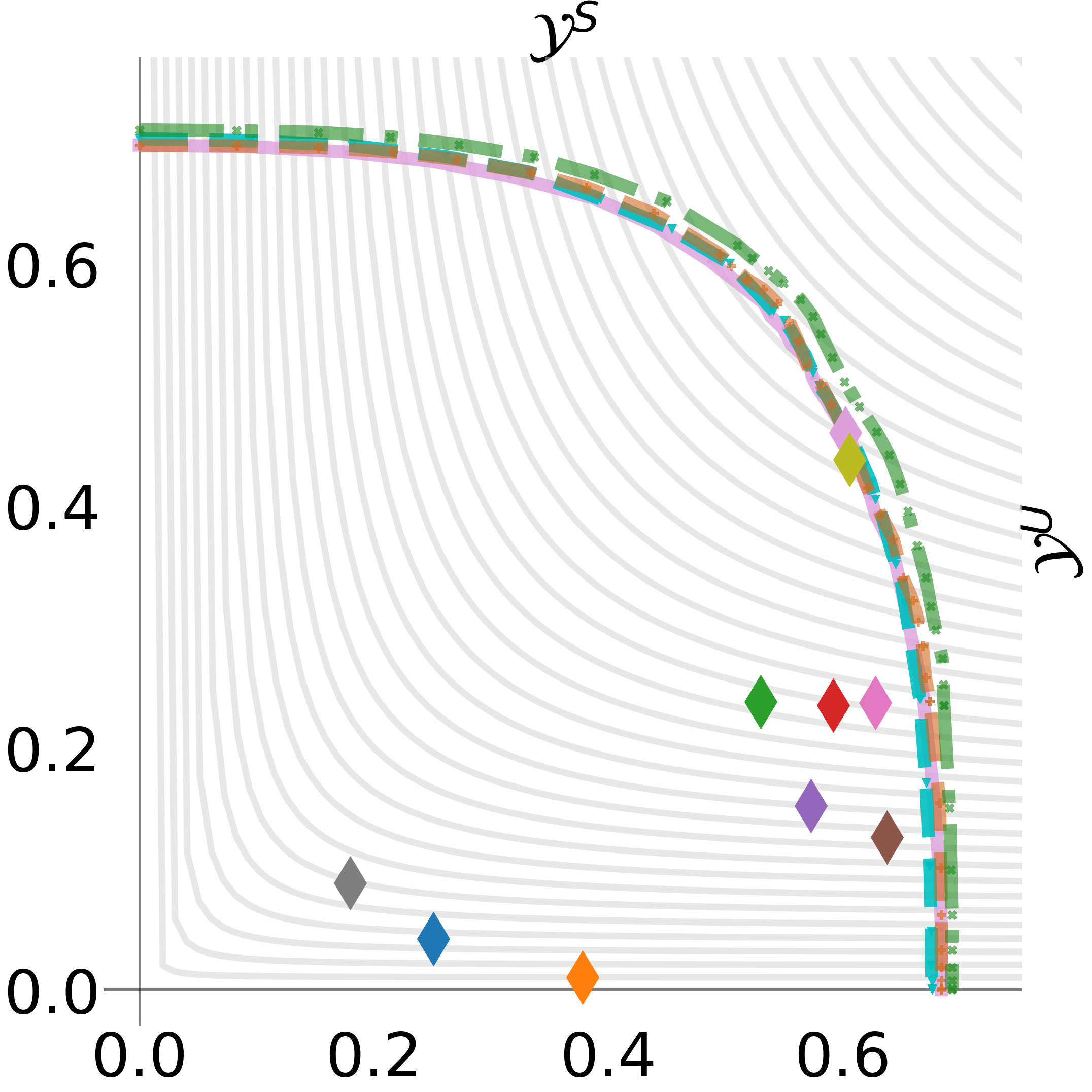}
     \caption{~CUB}
  \end{subfigure}
  \begin{subfigure}[b]{0.245\textwidth}
    \includegraphics[width=\textwidth, height=\textwidth]{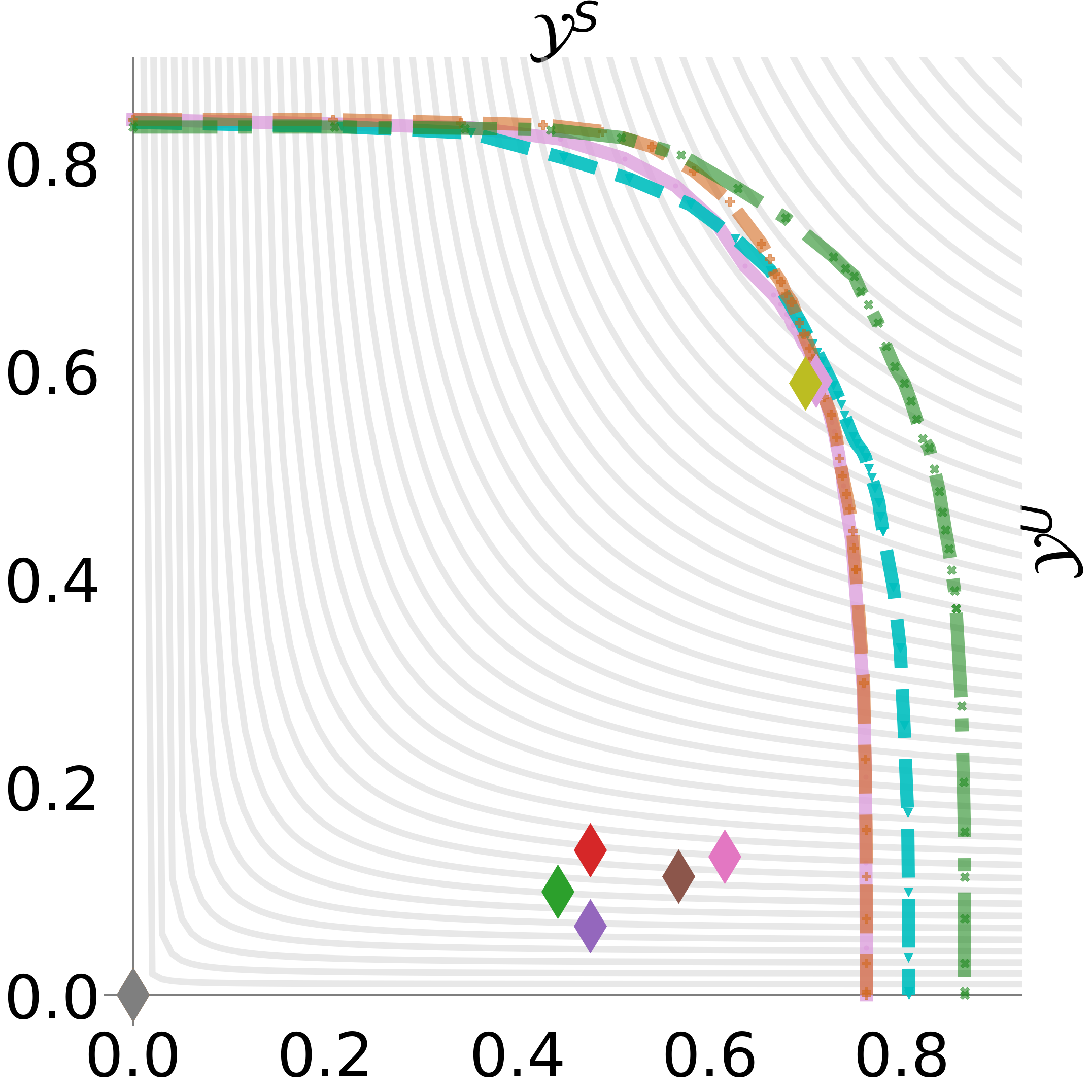}
     \caption{~FLO}
  \end{subfigure}
    \begin{subfigure}[b]{0.245\textwidth}
    \centering
    \includegraphics[width=\textwidth, height=\textwidth]{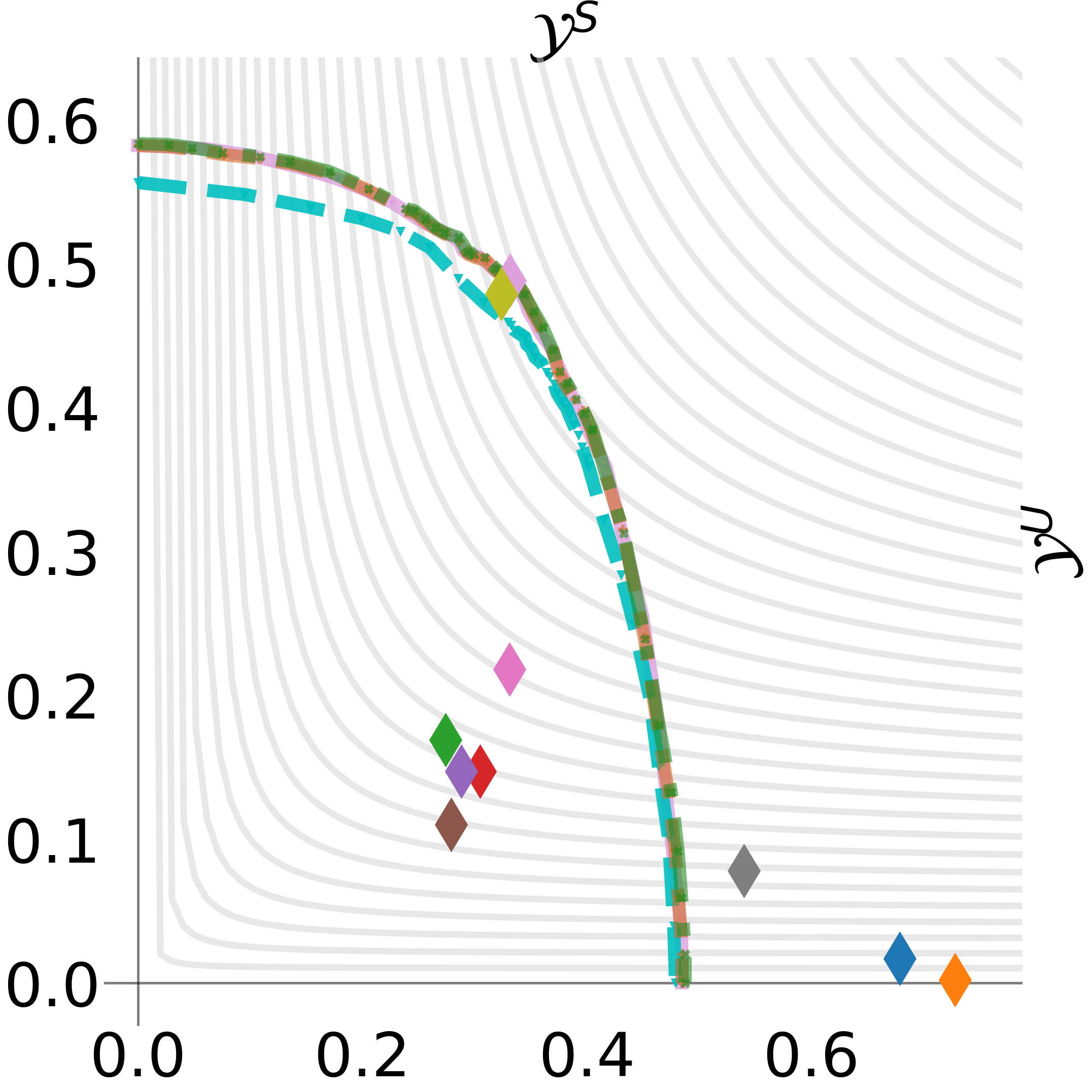}
     \caption{~SUN}
  \end{subfigure}
    \begin{subfigure}[b]{0.245\textwidth}
    \centering
    \includegraphics[width=\textwidth, height=\textwidth]{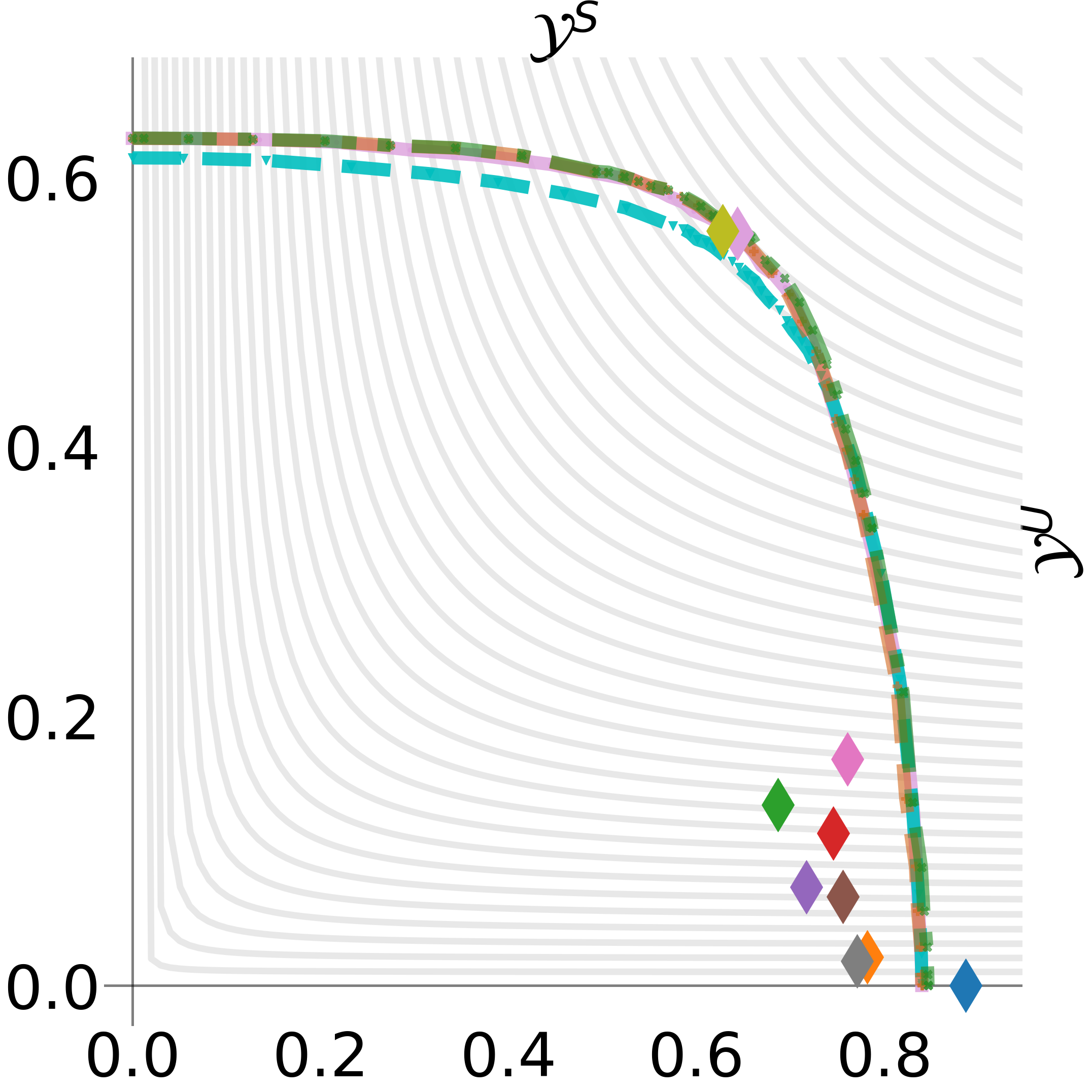}
     \caption{~AWA}
  \end{subfigure}
 
   \begin{subfigure}[b]{0.8\textwidth}
    \centering
    \includegraphics[width=\textwidth, height=0.15\textwidth]{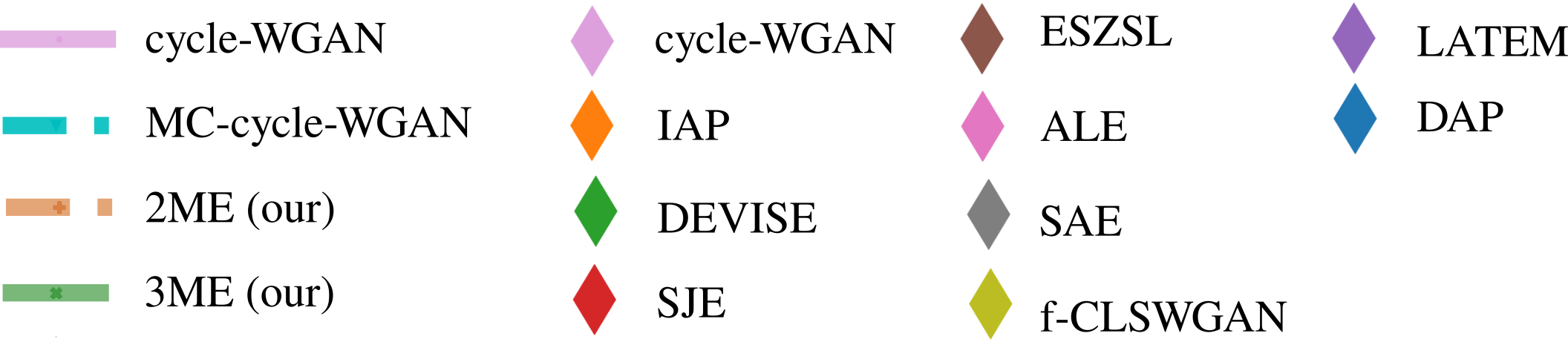}
\end{subfigure}
  \caption{Classification accuracy for seen and unseen classes for the proposed methods 2ME and 3ME, and several baseline and SOTA methods (please see text and Table~\ref{table:gzsl_results} for details about the methods and Fig.~\ref{fig:hp-search} for details about this graph).  Note that these graphs are used to compute the AUSUC in Table~\ref{table:gzsl_aucsuac}.}
  \label{fig:ausuac}
\end{figure*}

Figure~\ref{fig:ausuac} shows the seen and unseen classification graphs, similar to the ones presented in Fig.~\ref{fig:hp-search}, for previously published GZSL methods (please refer to Tab.~\ref{table:gzsl_results} for a reference to each method -- each method is represented by a diamond of a single color), and our proposed 2ME and 3ME.  In Figure~\ref{fig:ausuac}, we also show the curve for cycle-WGAN that extends the method in~\cite{felix2018multi} with the classification weighting (between seen and unseen classes) provided by~\eqref{eq:score-by-beta}, but without the MC dropout classification.  Moreover, we also show the curve for the MC-cycle-WGAN that extends the cycle-WGAN~\cite{felix2018multi} with the MC dropout.     
Note that we only report single points for previous methods~\cite{xian2017feature} because this is the result available from the literature (i.e., previous methods only report a single operating point for the classification of seen and unseen classes).  In principle, similar curves and the AUSUC can be computed for all other methods, but that is beyond the scope of this paper. 

Finally, using the graph in Fig.~\ref{fig:ausuac} we can compute the AUSUC on each dataset for cycle-WGAN, MC-cycle-WGAN, 2ME, and 3ME -- see Table~\ref{table:gzsl_aucsuac}. For all the models, we take the optimal value estimated for $\alpha$ (using the validation set, as explained above) and plot the seen and unseen classification curve by varying $\beta \in [0,1]$ in~\eqref{eq:score-by-beta}.  The AUSUC is the area under that curve.

\begin{table}
\centering
\caption{Area under the curve of seen and unseen accuracy (AUSUC).  The highlighted values per column represent the best results in each dataset.}
\label{table:gzsl_aucsuac}

    \begin{tabular}{|l|c|c|c|c|}
        \hline
        \textbf{Classifier}  
        & \textbf{CUB} 
        & \textbf{FLO} 
        & \textbf{SUN}
        & \textbf{AWA } \\
        \hline
        
        {cycle-WGAN} 
        & $0.418$ & $0.595$ & $0.235$ & $0.473$ \\
        
        {MC-cycle-WGAN} 
        & $0.420$ & $0.615$ & $0.225$ & $0.461$ \\
        
        {2ME} 
        & $0.423$ & $0.604$ & $0.235$ & $0.473$ \\
        
        \hline 
        {3ME} 
        & $\textbf{0.430}$ & $\textbf{0.673}$ & $\textbf{0.236}$ & $\textbf{0.477}$ \\
        \hline
     
    \end{tabular}
\end{table}


\section{Discussions}

\textbf{Harmonic mean.} In Table~\ref{table:gzsl_results}, we notice a clear tendency of our proposed model 3ME to perform substantially better in terms of H-mean on fine-grained datasets, such as CUB and FLO, and marginally better for coarse-grained datasets SUN and AWA. 
More specifically, on CUB we improve from $52.2\%$~\cite{felix2018multi} to $54.3\%$, on FLO from $64.5\%$~\cite{felix2018multi} to $66.8\%$, on SUN we notice a marginal improvement from $39.2\%$~\cite{felix2018multi} to $39.4\%$, and on AWA from $59.7\%$~\cite{felix2018multi} to $60.2\%$.
In terms of large-scale datasets, such as ImageNet, we show on Table~\ref{table:gzsl_imagenet} that our method outperforms the current SOTA in terms of the H-mean result.

\textbf{ZSL}. Although the scope of this paper it not the optimization of ZSL, we note that, when $\beta=1$, the proposed method outperforms the SOTA by a large margin on the datasets CUB (from $57.8\%$~\cite{felix2018multi} to $71.1\%$) and FLO (from $68.8\%$~\cite{felix2018multi} to $83.9\%$), and it maintains competitive results on SUN and AWA. Hence, based on these results we can argue that with our method, it is no longer necessary to build separate models for GZSL and ZSL.

\textbf{Seen and unseen classification graphs.} Figures~\ref{fig:hp-search} and~\ref{fig:ausuac} clearly show the trade-off between the classification of seen and unseen classes for GZSL methods. In particular, from Fig.~\ref{fig:ausuac}, it is interesting to notice a fact that is prevalent in GZSL methods, which is the classification imbalance that usually favors the seen classes -- it is clear from that figure that the majority of the SOTA methods (represented by diamonds) lie at the bottom-right part of the graphs, indicating the preference for seen classes.  
The optimization for $\alpha$ in \eqref{eq:alpha} and $\beta$ in \eqref{eq:score-by-beta} to maximize H-mean shows that the more balanced classification usually lies close to the elbow of the curve, located at the top-right part of the graph -- it is interesting to note that even though this optimization is performed with a validation set, it generalizes well to the test set, as shown in Fig.~\ref{fig:ausuac}.
\textbf{AUSUC.}  Table~\ref{table:gzsl_aucsuac} shows that the proposed approach 3ME outperforms all competing methods. In fact, the methods presented in this table shows an ablation study of the proposed method, where cycle-WGAN is a visual data augmentation method trained with classification weighting, MC-cycle-WGAN adds the MC dropout to the cycle-WGAN model, 2ME combines multi-modal ensemble with classification weighting, and 3ME consits of 2ME with MC dropout.  Note that we could also have added DAP with classification weighting and MC-DAP (classification weighting plus MC-dropout) to Table~\ref{table:gzsl_aucsuac}, but given the poor results of DAP in Fig.~\ref{fig:hp-search} (represented by $\alpha=0$), we decided to leave them out, so we would not clutter the result section. It is worth emphasizing that the AUSUC measure provides a more complete assessment of GZSL methods, where it is no longer necessary to commit to a particular operating point of the classification of seen and unseen classes. 

\section{Conclusions and Future Work}

In this paper, we introduce an new approach to perform GZSL by a multi-modal combination of ensemble classifiers using visual and semantic modalities.  Furthermore, we build a GZSL classifier that not only is optimized to produce a balanced classification of seen and unseen classes, but it can also work in various GZSL classification operating points (from ZSL to fully supervised classification).  This is performed by keeping the optimal value for $\alpha$ (responsible for combining the two modalities) and varying $\beta$ (responsible for the balance between the seen and unseen classification) without re-training any of the classifier parameters or hyper-parameters. 
The experimental results show that our proposed approach has the SOTA H-mean results for CUB, FLO, SUN, AWA, and Imagenet.  In particular, our results are substantially better than the SOTA on CUB and FLO, which are fine-grained datasets, and marginally better on SUN, AWA, and Imagenet, which are coarse-grained datasets. Furthermore, our model produces substantial improvements in terms of ZSL results for the fine-grained datasets CUB and FLO. Finally, the proposed approach also achieves the current SOTA results in terms of AUSUC for all the datasets.
In the future, we intend to study the reason behind the performance difference observed between fine-grained and coarse-grained datasets.  We are particularly interested in understanding why it is difficult to obtain high classification accuracy on the unseen classes of large scale coarse-grained datasets, such as ImageNet. In addition, we also plan to develop a more elegant end-to-end training approach that automatically learns $\alpha$ in~ \eqref{eq:alpha} and $\beta$ in~\eqref{eq:score-by-beta}.

\section{Acknowledgement}

Supported by Australian Research Council through grants DP180103232, CE140100016 and FL130100102.  We would like to acknowledge the donation of a TitanXp by Nvidia.


{\small
\bibliographystyle{ieee}
\bibliography{egbib}
}

\end{document}